%% file: 2022_NoyelBarbierFournelJourlin_DGMM.tex
\documentclass[runningheads]{llncs}
\input{header.tex}
\input{header_math_commands.tex}

\begin{document}
\title{Logarithmic Morphological Neural Nets robust to lighting variations\thanks{Supported by Lyon Informatics Federation, France, through the ``FakeNets'' project.}}
\titlerunning{Logarithmic Morphological Neural Nets}
\author{Guillaume Noyel\inst{1}\orcidID{0000-0002-7374-548X} \and
Emile Barbier-{}-Renard\inst{1}\orcidID{0000-0003-2967-6082} \and
Michel Jourlin\inst{1}\orcidID{0000-0002-2076-3465} \and 
Thierry Fournel\inst{1}\orcidID{0000-0002-1613-4594}}
\authorrunning{G. Noyel et al.}
\institute{Laboratoire Hubert Curien, UMR CNRS 5516, Universit\'e Jean Monnet, \num{42 000} Saint-Etienne, France \\
\email{\{guillaume.noyel,michel.jourlin,fournel\}@univ-st-etienne.fr}\\
\email{emile.barbier.renard@etu.univ-st-etienne.fr}}
\maketitle              % typeset the header of the contribution

\begin{abstract}
%The abstract should briefly summarize the contents of the paper in
%\textcolor{red}{To be written.}
Morphological neural networks allow to learn the weights of a structuring function knowing the desired output image. However, those networks are not intrinsically robust to lighting variations in images with an optical cause, such as a change of light intensity. In this paper, we introduce a morphological neural network which possesses such a robustness to lighting variations. It is based on the recent framework of Logarithmic Mathematical Morphology (LMM), i.e. Mathematical Morphology defined with the Logarithmic Image Processing (LIP) model. This model has a LIP additive law which simulates in images a variation of the light intensity. We especially learn the structuring function of a LMM operator robust to those variations, namely : the map of LIP-additive Asplund distances. Results in images show that our neural network verifies the required property.
\end{abstract}
%
%
%
%%%%%%%%%%%%%%%%%%%%%%%%%%%%%%%%%%%%%%%%%%%%%
%
%	Introduction
%
%%%%%%%%%%%%%%%%%%%%%%%%%%%%%%%%%%%%%%%%%%%%%
\section{Introduction}

Deep learning \cite{Goodfellow2016} based on convolutional neural networks (CNN) \cite{LeCun2015} has emerged as a methodology to learn a model of the data in order to perform a classification or a regression task \cite{Hastie2009}. %The model is composed of several layers of neurons. 
During the training phase, the model parameters %(i.e. the weights of the layers and of the convolution filters) 
are learnt by minimising a loss between a given truth and the model prediction.
%The loss minimisation is performed by an optimisation algorithm such as the stochastic gradient descent algorithm \cite{Goodfellow2016}.
%Convolutional neural nets (CNN) have brought a revolution in the computer vision community in terms of results \cite{LeCun2015}. 
In parallel to CNN, several morphological neural nets have been defined and studied. 
First, fully connected morphological neural nets (where the output depends on all the input pixels) have been defined in~\cite{Davidson1993} and more recently by Charisopoulos et al. \cite{Maragos2017}, Mondal et al. \cite{Mondal2019a} and Zhang et al. \cite{ZhangBlusseau2019}. 
Second, Barrera et al. defined morphological neural nets in sliding windows \cite{Barrera1997} (where the output only depends on the input pixels in the window). 
Moreover, deep morphological networks have also been defined by using either approximations of the morphological operations \cite{Masci2013,Mellouli2017,Mondal2019b,Saeedan2018,Shen2019,Kirszenberg2021,Aouad2022}, or exact morphological operations \cite{Franchi2020,Nogueira2021}.
Deep morphological networks have been used e.g. for classification in hyperspectral images \cite{Nogueira2021}, image de-hazing or de-raining \cite{Mondal2020}, image denoising \cite{Franchi2020}.
A morphological network has a constant additive invariance \cite{VelascoForero2022}. In classical neural networks, a CNN was designed to have a shift invariance \cite{Chaman2021} and a neural net has a contrast invariance based on quaternion local phase \cite{Moyasanchez2020}.

However, morphological neural networks are not intrinsically robust to real lighting variations. The analysis of images presenting such variations is a challenging task that can occur in many settings \cite{Jourlin2016,ZhangZhao2019,Noyel2020b}: industry, traffic control, underwater vision, face recognition, large public health databases, etc. 
In this paper, we propose a morphological neural network which is robust to such lighting variations due to a change of light intensity or of camera exposure-time.

Such a neural net is based on a metric, namely the functional Asplund metric \cite{Jourlin2016_chap3} which presents this robustness property. This metric is defined with the \textit{Logarithmic Image Processing} (LIP) model \cite{Jourlin2001,Jourlin2016} which models those lighting variations. As the LIP model is based on a famous optical law, namely the \textit{Transmittance Law}, we shall introduce in this way \textit{Physics} in those neural nets. 
In addition, the maps of Asplund distances between an image and a reference template, the probe, are related to Mathematical Morphology \cite{Noyel2020b}. We shall see that they are especially related to the newly introduced framework of \textit{Logarithmic Mathematical Morphology} \cite{Noyel2019a,Noyel2021}.

%%%%%%%%%%%%%%%%%%%%%%%%%%%%%%%%%%%%%%%%%%%%%%%%%%%%%%%%%%%%%%%%%%%%%%%%%%
%
%		Background
%
%%%%%%%%%%%%%%%%%%%%%%%%%%%%%%%%%%%%%%%%%%%%%%%%%%%%%%%%%%%%%%%%%%%%%%%%%%
\section{Background}
\label{sec:back}

%%%%%%--------------------------------------------------------------------
%		Logarithmic Image Processing
%%%%%%--------------------------------------------------------------------
\subsection{Logarithmic Image Processing}
\label{ssec:back:LIP}

The LIP model is defined for an image $f$ acquired by transmission and, as it is consistent with the human vision \cite{Brailean1991,Jourlin2016}, it can also be used for images acquired by reflection. In this model, the light is passing through a semi-transparent medium and is captured by the sensor. The resulting image $f$ is a function defined on a domain $D \subset \Real^2$ with values lying in the interval $\left[0,M\right[ \subset \Real$. It is important to note that the LIP greyscale is inverted with respect to the usual grey scale. $0$ corresponds to the white extremity, when all the light passes through the medium. $M$ is the black extremity, when no light is passing. For images digitised on 8-bits, $M$ is always equal to $2^8 = 256$. 

According to the \textit{transmittance law}, the transmittance $T_{f \LP g}$ of the superimposition $f \LP g$ of two media which generate the images $f$ and $g$, is equal to the product of the transmittances $T_f$ and $T_g$ of each image: $T_{f \LP g} = T_f \cdot T_g$. The transmittance $T_f$ of any medium generating the image $f$ is equal to $T_f = 1 - f / M$. From both previous equations, the LIP-addition law is deduced :
\begin{equation}
	f \LP g = f + g - f \cdot g/M. \label{eq:LIP:plus}%
\end{equation}
	As the addition $f \LP f$ may be written as $2 \LT f$, the LIP-multiplication of an image $f$ by a real number $\la$ is expressed as:
\begin{equation}
	\lambda \LT f = M - M \left( 1 - f/M \right)^{\lambda}. \label{eq:LIP:times}%
\end{equation}
When $\lambda = -1$, the LIP-negative function $\LM f = -1 \LT f$ can be defined, as well as the LIP-difference $f \LM g$ between two images $f$ and $g$. They are expressed as follows:
\begin{align}
	\LM f 	&= (-f)/(1-f/M), \label{eq:LIP:negative}\\
	f \LM g &= (f-g)/(1-g/M). \label{eq:LIP:minus}%
\end{align}
It can be noticed that $f \LM g$ is an image (i.e. $f \LM g\geq 0$) if and only if $f \geq g$. 

The LIP model has a \textit{strong mathematical property}. Let $\Fcurv_M = \left]-\infty,M\right[^D$ be the set of real functions defined on the domain $D$ and whose values are less or equal than $M$. Let $\I = \left[0,M\right[^D$ be the set of images. The set $(\Fcurv_M,\LP,\LT)$ is a \textit{real vector space} and the set $(\I,\LP,\LT)$ is its \textit{positive cone}. 

The LIP model also possesses a \textit{strong physical property}. The LIP-negative values $\LM f$, where $f\geq 0$, acts as light intensifiers. Those values can therefore be used to compensate the image attenuation in scenes captured with a low lighting. In particular, the LIP-addition of a positive constant to an image simulates the effect of a decrease of the light intensity or a decrease of the camera exposure-time. The resulting image is therefore darker than the original one. In an equivalent way, the LIP-subtraction of a positive constant from an image, simulates an increase and the resulting image becomes brighter.

%%%%%%--------------------------------------------------------------------
%		Logarithmic Mathematical Morphology
%%%%%%--------------------------------------------------------------------
\subsection{Logarithmic Mathematical Morphology}
\label{ssec:back:LMM}

Logarithmic Mathematical Morphology (LMM) was introduced by Noyel in \cite{Noyel2019a,Noyel2021}. LMM consists of defining morphological operations \cite{Serra1982,Heijmans1990} in the LIP framework. 
In LMM, the dilation $\delta_b^{\LP}$ and the erosion $\epsilon_b^{\LP}$ are defined in the lattice $(\Fcurv_M,\leq)$. 
Let $f$ and $b \in \Fcurvb_M$ be two functions, where $\Fcurvb_M = \left[-\infty,M\right]^D$.
The function $b : D \mapsto \left]-\infty,M\right[$ is chosen as the structuring function%, i.e. the function by which the function $f$ will be analysed.
, which implies that outside the domain $D_b \subset D$, all its values are equal to $-\infty$: $\forall x \in D \setminus D_b$, $b(x)=-\infty$.
Both mappings $\delta_b^{\LP}$ and $\varepsilon_b^{\LP}$ are named \textit{logarithmic-dilation} and \textit{logarithmic-erosion}, respectively. They are defined by:
\begin{align}
\delta_b^{\LP}(f)(x)		 &= \vee 	\left\{ f(x - h) \LP b(h), h \in D \right\} \label{eq:LIP-dilation}\\
\varepsilon_b^{\LP}(f)(x) &= \wedge \left\{ f(x + h) \LM b(h), h \in D \right\}. \label{eq:LIP-erosion}%
\end{align}
In the case of ambiguous expressions, the following conventions will be used: $f(x - h) \LP b(h) = -\infty$ when $f(x - h) = -\infty$ or $b(h) = -\infty$, and $f(x + h) \LM b(h) = M$ when $f(x + h) = M$ or $b(h) = -\infty$.

Both operations form an adjunction, i.e. for all $f,g \in \Fcurv_M$, $\delta_b^{\LP}
(g) \leq f \Leftrightarrow g \leq \varepsilon_b^{\LP}(f)$. As a consequence, the composition $\gamma_{b}^{\LP} = \delta_b^{\LP} \circ \varepsilon_b^{\LP}$ is an \textit{opening} and 
the composition $\varphi_{b}^{\LP} = \varepsilon_b^{\LP} \circ \delta_b^{\LP}$ is a \textit{closing}. LMM operations are adaptive to lighting variations modelled by the LIP-additive law. We shall see that at least an operation which is robust to those lighting variations can be defined in the LMM framework.

The logarithmic-dilation $\delta_b^{\LP}$ and the logarithmic-erosion $\varepsilon_b^{\LP}$, which are defined in the lattice $(\Fcurv_M,\leq)$, are related to the usual functional dilation $\delta_b$ (or $\oplus$) and erosion $\varepsilon_b$ (or $\ominus$), which are defined in the lattice of real functions $(\Real^D,\leq)$. These usual dilation and erosion are defined, for all $x \in D$, by :
\begin{align}
\delta_b(f)(x)		 &= (f \oplus b) (x) =\vee 	\left\{ f(x - h) + b(h), h \in D \right\}  \label{eq:dilate_funct}\\
\varepsilon_b(f)(x)   &= (f \ominus b) (x) = \wedge \left\{ f(x + h) - b(h), h \in D \right\} . \label{eq:erode_funct}%
\end{align}
Such relations are based on the isomorphism  $\xi: \Fcurvb_M \mapsto \Realb^D$ and its inverse  $\xi^{-1}: \Realb^D \mapsto \Fcurvb_M$, which are expressed by $\xi(f) = -M \ln{(1-f/M)}$ and $\xi^{-1}(f) = M(1-\exp{(-f/M)})$ \cite{Jourlin1995}.
The relations between the logarithmic-operations $\delta_b^{\LP}$, $\varepsilon_b^{\LP}$ and the usual functional erosion are the following ones~\cite{Noyel2019a,Noyel2021}:
	\begin{align}
		\delta_b^{\LP} (f) &= \xi^{-1} \left( \delta_{\xi(b)}[\xi(f)] \right) = \xi^{-1} \left[ \xi(f) \oplus \xi(b) \right] \label{eq:LIP_dil_prop}\\
%		&= M [1 - \exp{(- \delta_{\acute{b}}(\acute{f}) )}], \label{eq:LIP_dil_prop}\\
		\varepsilon_b^{\LP} (f) &= \xi^{-1} \left( \varepsilon_{\xi(b)} [\xi(f)] \right)  = \xi^{-1} \left[ \xi(f) \ominus \xi(b) \right], \label{eq:LIP_ero_prop}%\\
%		&= M [1 - \exp{(-\varepsilon_{\acute{b}}(\acute{f})   )}], \label{eq:LIP_ero_prop}%
	\end{align}
%	where $\acute{f}: D \mapsto \Realb$ is equal to $\acute{f}= -\ln{\left( 1 - f/M \right)}$.
where $f$ and $b \in \Real^D$.
These relations are not only important from a theoretical point of view but also from a practical point of view. Indeed, they facilitate the programming of the logarithmic operations by using the usual morphological operations which exist in numerous image analysis software.

\begin{figure}[hbt!]
\centering
\begin{subfigure}{0.49\textwidth}
\centering
\includegraphics[width=6cm]{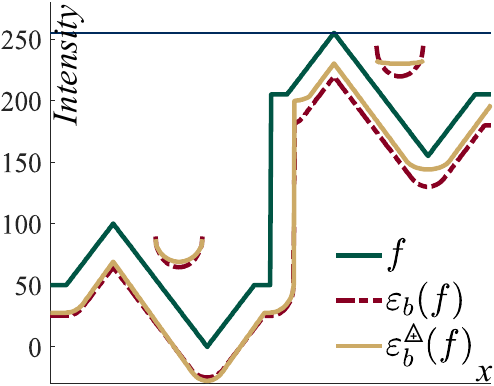}%
\caption{Erosions.}
\label{fig:LMM:comp_MM_vs_LMM_signal:1}
\end{subfigure}
\hfil
\begin{subfigure}{0.49\textwidth}
\centering
\includegraphics[width=6cm]{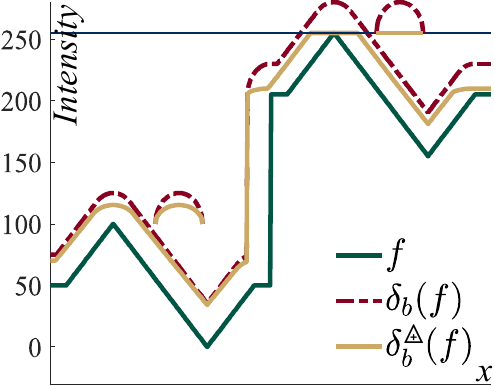}%
\caption{Dilations.}
\label{fig:LMM:comp_MM_vs_LMM_signal:2}
\end{subfigure}
%\\
%\begin{subfigure}{0.49\textwidth}
%\centering
%\includegraphics[width=6cm]{LIP_Opening_signal}%
%\caption{Ouvertures.}
%\label{fig:LMM:comp_MM_vs_LMM_signal:3}
%\end{subfigure}
%\begin{subfigure}{0.49\textwidth}
%\centering
%\includegraphics[width=6cm]{LIP_Closing_signal}%
%\caption{Fermetures.}
%\label{fig:LMM:comp_MM_vs_LMM_signal:4}
%\end{subfigure}
\caption{In an image $f$, comparison between functional MM and LMM for the 
(a) erosions $\varepsilon_b(f)$, $\varepsilon_b^{\protect \LP}(f)$ and 
(b) dilations $\delta_b(f)$, $\delta_b^{\protect \LP}(f)$. 
%(c) openings $\gamma_b(f)$, $\gamma_b^{\protect \LP}(f)$ and 
%(d) closings $\varphi_b(f)$, $\varphi_b^{\protect \LP}(f)$. 
(a) and (b) For both signal peaks, the structuring function $b$ is represented (after an horizontal translation) for the erosions $\varepsilon_b(f)$, $\varepsilon_b^{\protect \LP}(f)$ and the dilations $\delta_b(f)$, $\delta_b^{\protect \LP}(f)$.}
%\caption{Sur une image $f$, comparaison entre la MM fonctionnelle et la LMM pour les 
%(a) \'erosions $\varepsilon_b(f)$, $\varepsilon_b^{\protect \LP}(f)$, 
%(b) dilatations $\delta_b(f)$, $\delta_b^{\protect \LP}(f)$, 
%%(c) ouvertures $\gamma_b(f)$, $\gamma_b^{\protect \LP}(f)$ et 
%%(d) fermetures $\varphi_b(f)$, $\varphi_b^{\protect \LP}(f)$. 
%(a) et (b) Pour les deux pics du signal, la fonction structurante $b$ est repr\'esent\'ee (apr\`es une translation horizontale) pour les \'erosions $\varepsilon_b(f)$, $\varepsilon_b^{\protect \LP}(f)$ et les dilatations $\delta_b(f)$, $\delta_b^{\protect \LP}(f)$.}
\label{fig:LMM:comp_MM_vs_LMM_signal}
\end{figure}

In \figurename~\ref{fig:LMM:comp_MM_vs_LMM_signal}, in an image $f$, the logarithmic-erosion $\varepsilon_b^{\LP}(f)$ and dilation $\delta_b^{\LP}(f)$ are compared to the usual erosion $\varepsilon_b(f)$ and dilation $\delta_b(f)$. For the logarithmic operations, the amplitude of the structuring function $b$ varies according to the amplitude of the image $f$ ; whereas, for the usual morphological operations, the amplitude of $b$ does not change. In \figurename~\ref{fig:LMM:comp_MM_vs_LMM_signal:1}, when the image intensity is close to the maximal possible value $M=256$, the intensity of the logarithmic-dilation of $f$, $\delta_b^{\LP}(f)$, is less or equal than $M$, because the structuring function becomes flat ; whereas the intensity of the usual dilation of $f$, $\delta_b(f)$, is greater than $M$.

% ------------------- Figure comparaison sur un signal de la LMM avec la MM

%%%%%%--------------------------------------------------------------------
%		The LIP-additive Asplund's metric
%%%%%%--------------------------------------------------------------------
\subsection{The LIP-additive Asplund metric}
\label{ssec:back:LIPAddAsp}

%articles~\cite{Asplund1960}

The LIP-additive Asplund metric was defined by
 Jourlin~\cite{Jourlin2016_chap3}. 
\begin{definition}[LIP-additive Asplund metric]
 Let $f$ and $g \in \Fcurv_M $ be two functions. One of them, e.g. $g$, is chosen as a probing function. Both following numbers are then defined by: $c_1 = \inf{ \{c, f \leq c \LP g \}}$ and $c_2 = \sup{ \{c, c \LP g \leq f \} }$, where $c$ lies within the interval $]-\infty,M[$. %$c_1$ and $c_2$ are the constants to be LIP-added to the probe $b$ such that it is in contact with the function $f$ from above or from below, respectively. 
 The LIP-additive Asplund metric $d_{asp}^{\protect \LP}$ is defined by %$d_{asp}^{\LP}(f,g) = c_1 \LM c_2$.
\begin{equation}
	d_{asp}^{\LP}(f,g) = c_1 \LM c_2. \label{eq:dasAdd}%
\end{equation}
\end{definition}

\begin{property}[Robustness to lighting variations~\cite{Jourlin2016_chap3}]
Importantly, this metric is invariant under lighting changes modelled by a LIP-addition of a constant:
\begin{equation}
\forall k \in ]-\infty,M[,\quad d_{asp}^{\LP}(f,g) = d_{asp}^{\LP}(f \LP k , g) \text{ and } d_{asp}^{\LP}(f,f \LP k) = 0.
\label{eq:dasAdd_rob_LIPadd} 
\end{equation}
Those changes correspond to a modification of the light intensity or of the camera exposure-time.
\end{property}

%%%%%%--------------------------------------------------------------------
%		Learning the structuring function in morphological operations
%%%%%%--------------------------------------------------------------------
\subsection{Learning of the structuring function in morphological operations}
\label{ssec:back:LearnSF_MM}

For machine learning in Mathematical Morphology (MM), the functions are defined in discrete grids of $\Zint^2$ ;  $f : D \mapsto \Real$ and  $b : D_b \mapsto \Real$, where $D_b \subset D \subset \Zint^2$. Let the cardinal of the set $D_b$ be equal to $2n+1$. The bidimensional functions can always be represented as unidimensional arrays, e.g. by concatenating their rows. The structuring function are therefore written as follows: $b = \{b_{-n}, \ldots , b_{n}\}$. 
The dilation $\delta_b$ and the erosion $\varepsilon_b$ layers can be expressed by \cite{Franchi2020,Mondal2020}:
\begin{align}
\delta_b(f)(x)		 &= \max_{i \in [\![-n , n]\!]} 	\left\{ f(x - i) + b_i\right\}  \label{eq:dilate_num}\\
\varepsilon_b(f)(x)   &= \min_{i \in [\![-n , n]\!]} \left\{ f(x + i) - b_i \right\} . \label{eq:erode_num}%
\end{align}
An image $f$ is passed through a dilation layer or an erosion layer that gives an output equal to $\widehat{g}$ (i.e. $\widehat{g} = \delta_b(f)$ or $\widehat{g} = \varepsilon_b(f)$). The weights of the structuring function $b$ are learnt so that the \textit{loss} $L(b) = L(g,\widehat{g})$  between the output $\widehat{g}$ of the neural net layer and the desired output $g$, is minimised.
%the output $\widehat{g}$ is as close as possible to the desired output $g$. Let $L$ be the \textit{loss} or the \textit{objective function} between the output $\widehat{g}$ of the neural net layer and the desired output $g$, i.e.,
%\begin{equation}
%L(b) = L(g,\widehat{g})\label{eq:obj_fct}.
%\end{equation}
%When the weights of the structuring function are learnt, then $g \rightarrow \widehat{g}$ and $L(b) \rightarrow 0$.
%Those weights are learnt by minimising the loss thanks to a \textit{stochastic gradient descent} algorithm \cite{Goodfellow2016}. This algorithm requires to compute the contribution (i.e. the derivative) of each weight to the loss. This is the \textit{back-propagation} \cite{LeCun2015} which is performed by using the \textit{chain rule} due to partial derivatives:
The loss minimisation is performed by a \textit{stochastic gradient descent} algorithm \cite{Goodfellow2016}, which requires to compute the derivative of each weight to the loss:
\begin{align}
\frac{\partial L}{\partial b_i} &= \sum_x \frac{\partial \widehat{g}(x)}{\partial b_i} \frac{\partial L}{\partial \widehat{g}(x)} = \sum_x \nabla \widehat{g}(x) \frac{\partial L}{\partial \widehat{g}(x)}.
\label{eq:chain_rule}
\end{align}
Let us denote $i^{x_*} = \arg \max_{i \in [\![-n , n]\!]}\{f(x-i) + b_i\}$ or $i^{x_*} = \arg \min_{i \in [\![-n , n]\!]}\{f(x+i) - b_i\}$, the index for which the dilation or the erosion takes its value. For the dilation, the gradient $\nabla \widehat{g}$ is equal to :
\begin{equation}
\nabla \widehat{g}(x) = 
\begin{cases}
	1  & \text{if } \widehat{g}(x) = f(x - i^{x_*}) + b_{i^{x*}}\\
    0, & \text{otherwise}
\end{cases}
\label{eq:dil_nn_gradient}
\end{equation}
and for the erosion to 
\begin{equation}
\nabla \widehat{g}(x) = 
\begin{cases}
	-1  & \text{if } \widehat{g}(x) = f(x + i^{x_*}) - b_{i^{x*}}\\
    0, & \text{otherwise.}
\end{cases}
\label{eq:ero_nn_gradient}
\end{equation}
The structuring function is therefore updated as : $b(x) = b(x) - \alpha \partial L / \partial b(x)$.

%%%%%%%%%%%%%%%%%%%%%%%%%%%%%%%%%%%%%%%%%%%%%%%%%%%%%%%%%%%%%%%%%%%%%%%%%%
%
%		The map of LIP-additive Asplund distances and LMM
%
%%%%%%%%%%%%%%%%%%%%%%%%%%%%%%%%%%%%%%%%%%%%%%%%%%%%%%%%%%%%%%%%%%%%%%%%%%
\section{Maps of LIP-additive Asplund distances}
\label{sec:MapLIPAddAsp}

%%%%%%--------------------------------------------------------------------
%		Definition
%%%%%%--------------------------------------------------------------------
\subsection{Definition}
\label{ssec:MapLIPAddAsp:def}

\begin{definition}[Map of LIP-additive Asplund distances]
Let $f \in \Fcurv_M$ be a function and $b \in \left]-\infty,M\right[^{D_{b}}$ a probe, where $D_b \subset D$. The map of Asplund distances is the mapping $Asp_{b}^{\LP}: \Fcurv_M \mapsto \I$ defined by:
\begin{align}
	Asp_{b}^{\LP} f(x) &= d^{\LP}_{asp} (f_{\left|D_b(x)\right.},b), \label{eq:map_As_add}%
\end{align}
\label{def:map_As_add}
\end{definition}
where $f_{\left|D_b(x)\right.}$ is the restriction of $f$ to the neighbourhood $D_b(x)$ centred on $x \in D$. The LIP addition $\LP$ makes the map of distances robust to contrast variations due to exposure-time changes: $\forall c \in \left]-\infty,M\right[$\,, $Asp_{b}^{\LP} (f \LP c) = Asp_{b}^{\LP} (f)$.

%%%%%%--------------------------------------------------------------------
%		Link with Logarithmic Mathematical Morphology
%%%%%%--------------------------------------------------------------------
\subsection{Link with Logarithmic Mathematical Morphology}
\label{ssec:MapLIPAddAsp:LMM}

The map of Asplund distances is related to Mathematical Morphology (MM) \cite{Noyel2017a,Noyel2020b}. Noyel as shown in \cite{Noyel2021} that it is specifically related to LMM as follows.

\begin{proposition}
	Let $f \in \Fcurvb_M$ be a function and $b \in \Fcurvb_M$ be a structuring function, where for all $x \in D_b$, $D_b \subset D$, $b(x) > -\infty$. The map of Asplund distances between the function $f$ and the structuring function $b$ is equal to:
\begin{align}
	Asp_{b}^{\LP} f &= \delta_{\LM \overline{b}}^{\LP} (f) \LM \varepsilon_b^{\LP}(f). \label{eq:map_AsAdd_LMM}%
\end{align}
%	For the \textit{mlub} and the \textit{mglb} of $f$, $c_{1_b} f$ and $c_{2_b} f$ we have:
%\begin{align}
%c_{1_b} f &= \delta_{\LM \overline{b}}^{\LP} (f),\label{eq:mlub_LMM}\\
%c_{2_b} f &= \varepsilon_b^{\LP}(f). \label{eq:mglb_LMM}%
%\end{align}
$\overline{b}$ is the reflected structuring function defined by $\forall x \in \overline{D}_b$, $\overline{b}(x)=b(-x)$.
In the case of ambiguous expressions, the following conventions will be used: $Asp_{b}^{\LP} f(x) = M$ when $\delta_{\LM \overline{b}}^{\LP} (f)(x) = M$ or $\varepsilon_b^{\LP}(f)(x) = -\infty$, and $Asp_{b}^{\LP} f(x) = 0$ when $\delta_{\LM \overline{b}}^{\LP} (f)(x) = \varepsilon_b^{\LP}(f)(x)$. 
	\label{prop:link_AsAdd_LMM}
\end{proposition}

%%%%%%%%%%%%%%%%%%%%%%%%%%%%%%%%%%%%%%%%%%%%%%%%%%%%%%%%%%%%%%%%%%%%%%%%%%
%
%		Map of Asplund distances Neural nets
%
%%%%%%%%%%%%%%%%%%%%%%%%%%%%%%%%%%%%%%%%%%%%%%%%%%%%%%%%%%%%%%%%%%%%%%%%%%
\section{Neural net of map of Asplund distances}
\label{sec:MapLIPAddAspNN}

%%%%%%%--------------------------------------------------------------------
%%		Learning the structuring function for the Map of Asplund distances
%%%%%%%--------------------------------------------------------------------
%\subsection{Learning the structuring function for the Map of Asplund distances}
%\label{ssec:MapLIPAddAsp:LearnSF}

From equations~\eqref{eq:map_AsAdd_LMM}, \eqref{eq:LIP_dil_prop}, \eqref{eq:LIP_ero_prop} and knowing that $\xi(\LM b)= -\xi(b)$ and $\xi(f \LM g) = \xi(f) - \xi(g)$, one deduces that:
\begin{align}
	Asp_{b}^{\LP} f &= \xi^{-1}\left[ \delta_{ -\xi(\overline{b})} \xi(f) - \varepsilon_{\xi(b)} \xi(f) \right]. \label{eq:map_AsAdd_LMM_neural_net}%
\end{align}
We then create a \textit{map of Asplund distance layer} $Asp_{b}^{\LP}$,  where we apply this layer to the input image $f$ in order to give an output $\widehat{g}=Asp_{b}^{\LP}(f)$. The structuring function $b$ is learnt so as to minimise a loss $L(b)=L(g,\widehat{g})$ between $\widehat{g}$ and a desired output $g = Asp_{b_r}^{\LP}(f)$, where $b_r$ is a reference structuring function. The goal is to learning $b$ in order to discover $b_r$.

%In practice, $b$ is initialised as a null real matrix of size $A\times B$ pixels.

In morphological neuron implementations \cite{Franchi2020}, learning $b$ is equivalent to learn the weight matrix $W\in\mathbb{R}^{A\times B}$, where $W(x)=b(x)$, for all $x\in D_b$ and $W(x)=-\infty$,  otherwise. $A\times B$ is the window size in pixels of $D_{W} \subset D$. However, in equation~\eqref{eq:map_AsAdd_LMM}, in the term $\delta_{\LM \overline{b}}^{\LP}$, we also expect that for all $x\notin D_b$, $\LM \overline{W}(x) = -\infty$, which is not compatible with the current definition of $W$.

We therefore introduce a definition of $b$ relying on two learnt kernels: the height kernel $W_h$ and the mask kernel $W_m \in \mathbb{R}^{A\times B}$. First, $W_h$ corresponds to the height-map of the probe satisfying $W_h(x)=b(x)$, for all $x\in D_b$. Second, $W_m$ characterises the definition domain of the probe $D_b=\left\{x \in D_{W_m} \mid W_m(x)>0\right\}$. 
We then rewrite equation~\eqref{eq:map_AsAdd_LMM_neural_net} as:
\begin{equation}
	Asp_{b}^{\LP} f = \xi^{-1} \left[ \delta_{b_{dil}}(\xi(f))-\varepsilon_{b_{ero}}(\xi(f)) \right], \label{eq:map_AsAdd_LMM_neural_net_impl1}%
\end{equation}
where, $\forall x\in D_{W_m}$:
\begin{align}
	b_{dil}(x) &= 
	\begin{cases}
	-\xi(\overline{W_h})(x) & \text{if } \overline{W_m}(x)>0\\
	-\infty & \text{otherwise}
	\end{cases},\label{eq:map_AsAdd_LMM_neural_net_formal_dilation}%\\
	b_{ero}(x) = 
	\begin{cases}
	\xi(W_h)(x) & \text{if } W_m(x)>0\\
	-\infty & \text{otherwise}\\
	\end{cases}. %\label{eq:map_AsAdd_LMM_neural_net_formal_erosion}
\end{align}
In order to ensure that the gradient descent is smooth, a soft-binarisation function $\chi:\mathbb{R}^{A\times B} \mapsto ]0,1[^{A\times B}$, such as the sigmoid $\chi(v)=1/(1+\exp{(-v)})$, is applied to the mask kernel $W_m$ instead of a threshold. We therefore define $V = \chi(W_m)\in \left]0,1\right[^{A\times B}$ as the soft-mask of the probe in the window of size $A\times B$. Because $V$ is not a binary mask, $-\infty$ cannot be used when computing $b_{ero}$ or $b_{dil}$. As $f$ is an image, we have for all $x\in D$, $f(x)\in \left[0,M-1\right]$. This implies that $\xi(f(x)) \in \left[0,\xi(M-1)\right]$. A bottom value $\bot = -\xi(M-1)$ is chosen such that $\xi(f)(x)-\bot \geq \xi(M-1)$. This implies that $\xi(f)(x)+\bot \geq \xi(0)$. We then define the approximations $\tilde{b}_{dil}$ and $\tilde{b}_{ero}$ of $b_{dil}$ and $b_{ero}$ as follows:
\begin{align}
	\tilde{b}_{dil} &= -\xi(\overline{W_h})\cdot\overline{V}+\bot\cdot(1-\overline{V})\label{eq:map_AsAdd_LMM_neural_net_dilation_probe}\\
	\tilde{b}_{ero} &= \xi(W_h)\cdot V + \bot\cdot(1-\overline{V}).\label{eq:map_AsAdd_LMM_neural_net_erosion_probe}
\end{align}
From equations~\eqref{eq:map_AsAdd_LMM_neural_net_impl1}, \eqref{eq:map_AsAdd_LMM_neural_net_dilation_probe}, \eqref{eq:map_AsAdd_LMM_neural_net_erosion_probe}, an expression of $\widehat{g}$ is deduced: 
\begin{align}
	\widehat{g} = \xi^{-1} \left[ \delta_{\tilde{b}_{dil}}(\xi(f))-\varepsilon_{\tilde{b}_{ero}}(\xi(f)) \right]. \label{eq:map_AsAdd_LMM_neural_net_impl}%
\end{align}
It only contains components with derivatives and which can be used in the back-propagation algorithm.
In practice, $W_h$ and $W_m$ are both initialised as null matrices. 
\begin{remark}
In order to push the weights of the kernel $W_h$ away from zero, one might introduce a mean Gaussian or a mean squared Gaussian, as a regularisation function of $W_m$. This idea will be explored in future  works.
\end{remark}

%%%%%%%%%%%%%%%%%%%%%%%%%%%%%%%%%%%%%%%%%%%%%%%%%%%%%%%%%%%%%%%%%%%%%%%%%%
%
%		Illustration
%
%%%%%%%%%%%%%%%%%%%%%%%%%%%%%%%%%%%%%%%%%%%%%%%%%%%%%%%%%%%%%%%%%%%%%%%%%%
\section{Illustration and results}
\label{sec:ill}

We have illustrated our LMM network by using the Fashion MNIST dataset composed of a training set of \num{60 000} images and a test set of \num{10 000} images \cite{FashionMNIST}. Each image is digitised with a 8-bit greyscale and has a size of $\num{28}\times\num{28}$ pixels.

The goal was to learn a structuring function (or probe) $b$ (represented by both matrices $W_h$ and $W_m$) % which are initialised with null values) 
so as to discover a reference structuring function $b_r$.
This reference structuring function was defined as follows, for all $x \in D_{W_{b_r}}$, where $D_{W_{b_r}} \subset D$ and $W_{b_r}$ is a matrix of size $7 \times 7$:
\begin{equation}
b_r(x) = 
\begin{cases}
h(x) & \text{if }  h(x) \geq 0\text{, where } h(x) = -\beta \sqrt{7} \Vert x \Vert^2  \LP c\\
-\infty &  \text{otherwise.}
\end{cases}
\label{eq:struct_fct_ref}%
\end{equation}
Let $D_{b_r}= \{ x \in D_{W_{b_r}} \mid h(x) \geq 0\}$ be the domain of the probe.
The mask kernel $W_{m,r}$ and the height kernel $W_{h,r}$ of the reference structuring function $b_r$ are defined by  $W_{m,r} = \Ind_{D_{b_r}}$ and $\forall x \in D_{W_{b_r}}$, $W_{h,r}(x) = b_r(x)$, if $x \in D_{b_r}$ and $W_{h,r}(x) = 0$, otherwise. $\Ind_{D_{b_r}}$ is the indicator function of the set $D_{b_r}$. % defined by $\Ind_{D_b}(x)=1$, if $x \in D_b$ and $\Ind_{D_b}(x)=0$ otherwise.

By varying the parameters $\beta \in \{\num{0.2}, \num{0.4}, \ldots , \num{1.2}\}$ and $c \in \{\num{10}, \num{25}, \ldots , \num{250}\}$, a total of \num{102} reference structuring functions $b_r$ were generated.
With those $b_r$, the desired outputs $g = Asp_{b_r}^{\LP}(f)$ (i.e. a ground-truth) were computed in both train and test datasets. 
In the train set, the weights of $b$ were learnt by minimising a loss $L(g,\widehat{g})$. %$b$ was initialised as a null matrix of the same size as $b_r$. 
In the test set, the \textit{map of Asplund distance layer} $\widehat{g} = Asp_{b}^{\LP}(f)$ was applied to the images $f$  with the learnt structuring function $b$. The average of a validation metric $Val_m$ was computed between the estimated outputs $\widehat{g}$ and the ground-truth $g$.
For the loss $L$ and the validation metric $Val_m$, we used the mean square error $MSE$ or the LIP-mean square error $LIPMSE$ \cite{Pinoli1992}:
\begin{align}
MSE(g,\widehat{g}) &= \frac{1}{P}\sum_{i=1}^P \left[g_i - \widehat{g}_i \right]^2\label{eq:MSE}\\
LIPMSE(g,\widehat{g}) &= \frac{M^2}{P}\sum_{i=1}^P \left[\ln{\left(\frac{M - g_i}{M - \widehat{g}_i}\right)} \right]^2,\label{eq:LIPMSE}
\end{align}
where $P$ is the number of pixels of $g$.
The results of the validation metrics are shown in table~\ref{tab:res:test}.
In order to verify the robustness to lighting variations of our neural network, we have performed two other experiments with the same train set, but two additional test sets. We had therefore three test sets.  
\begin{inparaenum}[i)]
\item The first test set is the initial test set.
\item The  second test set is composed of the images of the first test set which were darkened by LIP-adding to them a constant of \num{100}. % grey levels. 
\item The third test set is composed of the images of the first test set which were brightened by LIP-subtracting from them a constant of \num{100}.
\end{inparaenum}
In table~\ref{tab:res:test}, one can notice that the averaged validation metrics between the three test sets are similar with a residual difference less than $\num{1.2e-06}$ grey levels. % and by $\num{1e-05}$ for the standard deviation.
Our neural network is therefore robust to lighting variations which are modelled by the LIP-addition of a constant. 

%\begin{table}
%\centering
%\caption{Comparison of the validations metrics $Val_m$ in three test sets: average, standard deviation and absolute average differences with the 1\textsuperscript{st} test set. Parameters: \num{15} epochs, Adam optimiser, learning rate $\alpha =0.5$, batch size: \num{20}.}
%\begin{tabular}{l@{\hspace{1em}}c@{\hspace{2em}}
%S[table-number-alignment = center,table-format=1.5e-03]
%S[table-number-alignment = center,table-format=1.5e-03]
%S[table-number-alignment = center,table-format=0.1e-03]}
%\hline
%Test sets 						& Metrics &\mcc{Averages} & \mcc{Std dev.} & \mcc{Abs. av. diff.}\\
%\hline
%1\textsuperscript{st} test set 	& $MSE$ 		& 9.74008e-05 & 0.67323e-03 & \mcc{}\\
%(initial, $f$) 					& $LIPMSE$ 	& 6.52817e-04 & 2.58031e-03 & \mcc{}    \\
%\hline
%2\textsuperscript{nd} test set 	& $MSE$ 		& 9.74008e-05 & 0.67323e-03 & 0\\ 
%(dark, $f \LP 100$)  		& $LIPMSE$ 	& 6.52887e-04 & 2.58048e-03  & 7e-08       \\
%\hline
%3\textsuperscript{rd} test set 	& $MSE$ 		& 9.74009e-05 & 0.67329e-03 & 1e-10\\
%(bright, $f \LM 100$)  		& $LIPMSE$ 	& 6.51606e-04 & 2.5757e-03  & 1.2e-06 \\
%\hline
%\end{tabular}
%\label{tab:res:test}
%\end{table}

\begin{table}
\centering
\caption{Comparison of the validations metrics $Val_m$ in three test sets: average, standard deviation and absolute average differences with the ground truth of the 1\textsuperscript{st} test set. Parameters: \num{15} epochs, Adam optimiser, learning rate $\alpha =0.5$, batch size: \num{20}.}
\begin{tabular}{l@{\hspace{1em}}c@{\hspace{2em}}
S[table-number-alignment = center,table-format=1.5e-03]
S[table-number-alignment = center,table-format=1.5e-03]
S[table-number-alignment = center,table-format=0.1e-03]}
\hline
Test sets 						& Metrics &\mcc{Averages} & \mcc{Std dev.} & \mcc{Abs. av. diff.}\\
\hline
1\textsuperscript{st} test set 	& $MSE$ 		& 9.740e-05 & 0.673e-03 & \mcc{}\\
(initial, $f$) 					& $LIPMSE$ 	& 6.528e-04 & 2.580e-03 & \mcc{}    \\
\hline
2\textsuperscript{nd} test set 	& $MSE$ 		& 9.740e-05 & 0.673e-03 & 0\\ 
(dark, $f \LP 100$)  			& $LIPMSE$ 	& 6.529e-04 & 2.581e-03 & 7e-08       \\
\hline
3\textsuperscript{rd} test set 	& $MSE$ 		& 9.740e-05 & 0.673e-03 & 1e-10\\
(bright, $f \LM 100$)  			& $LIPMSE$ 	& 6.516e-04 & 2.576e-03 & 1.2e-06 \\
\hline
\end{tabular}
\label{tab:res:test}
\end{table}

The error $E_{pr}(W_h,W_{h,r})$ between the height kernels of the learnt probe $b$ and of the reference probe $b_r$ is defined as follows: 
\begin{equation}
    E_{pr}(W_h,W_{h,r}) = \frac{1}{A.B} \left[ \min_{k}\sum_{x\in D_{b_r}} \left(W_{h,r}(x)-\left(W_{h}(x) \LP k \right)\right)^2 + \sum_{x\notin D_{b_r}} W_{h}^2(x)\right]\label{eq:mse_probes}
\end{equation}
It takes into account that the map of Asplund distances is invariant under the LIP-addition of a constant to its probe. %The second term penalises the non null weights outside the probe mask $D_{b_r}$.
%Where $k$ represents the LIP-shift of the probe, and $b_{rec}$ the reconstructed probe defined for all $x\in D_{W_h}$ as:
%\begin{equation}
%b_{rec}(x)=
%\begin{cases}
%	W_h(x)  & \text{if } W_m(x)>0\\
%    0, & \text{otherwise}
%\end{cases}
%\label{eq:reconstructed_probe}
%\end{equation}    
Table~\ref{tab:res:sf} shows the average errors between the kernels $W_m$ and $W_h$ of the learnt structuring function $b$ and the kernels $W_{m,r}$ and $W_{h,r}$ of the reference structuring function $b_r$, in the train set. %$\Ind_{D_{b_r}}$ is the indicator function of the set $D_{b_r}$. % defined by $\Ind_{D_b}(x)=1$, if $x \in D_b$ and $\Ind_{D_b}(x)=0$ otherwise.
One can notice that the average MSE between the soft mask kernels $W_m$ and $W_{m,r}$ is very small, with a value in $\num{1e-05}$.
The average error between the height kernels $W_h$ and $W_{h,r}$ has a value in \num{1e-04} grey levels. This means that the learnt kernels $W_m$ and $W_h$ are similar to the reference kernels $W_{m,r}$ and $W_{h,r}$.
The learnt probe $b$
is therefore similar to the reference probe $b_r$. 

\begin{table}
\centering
\caption{In the train set, comparison of the average errors between the kernels $W_m$ and $W_h$ of the learnt probe $b$ and the kernels $W_{m,r}$, $W_{h,r}$ of the reference probe $b_r$. A total of \num{102} probes $b$ have been learnt.}
\begin{tabular}{l@{\hspace{1em}}c@{\hspace{2em}}
S[table-number-alignment = center,table-format=1.2e-03]
S[table-number-alignment = center,table-format=1.2e-03]}
\hline
Kernels						& Errors &\mcc{Averages} & \mcc{Std dev.}\\
\hline
Heights of the probe $W_h$ 	& $E_{pr}$ 		& 6.89e-05 & 4.50e-04\\
Soft mask of the probe $W_m$ 	& $MSE$ 	& 2.23e-04 & 7.74e-04 \\
\hline
\end{tabular}
\label{tab:res:sf}
\end{table}

\begin{figure}[hbt!]
\centering
\begin{subfigure}{0.24\textwidth}
\centering
\includegraphics[width=2cm]{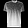}%
\caption{$f$}
\label{fig:FashionMNIST:1}
\end{subfigure}
\hfil
\begin{subfigure}{0.24\textwidth}
\centering
\includegraphics[width=2cm]{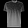}%
\caption{$f \LP 100$}
\label{fig:FashionMNIST:2}
\end{subfigure}
\hfil
\begin{subfigure}{0.24\textwidth}
\centering
\includegraphics[width=2cm]{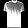}%
\caption{$f \LM 100$}
\label{fig:FashionMNIST:3}
\end{subfigure}
\hfil
\begin{subfigure}{0.24\textwidth}
\centering
\includegraphics[width=2cm]{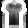}%
\caption{$g = Asp_{b_r}^{\protect \LP}f$}
\label{fig:FashionMNIST:4}
\end{subfigure}
\\
\begin{subfigure}{0.24\textwidth}
\centering
\includegraphics[width=2cm]{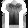}%
\caption{$Asp_{b}^{\protect \LP}f$\\\hfil}
\label{fig:FashionMNIST:5}
\end{subfigure}
\hfil
\begin{subfigure}{0.24\textwidth}
\centering
\includegraphics[width=2cm]{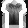}%
\caption{$Asp_{b}^{\protect \LP}(f \protect \LP 100)$}
\label{fig:FashionMNIST:6}
\end{subfigure}
\hfil
\begin{subfigure}{0.24\textwidth}
\centering
\includegraphics[width=2cm]{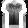}%
\caption{$Asp_{b}^{\protect \LP}(f \protect \LM 100)$}
\label{fig:FashionMNIST:7}
\end{subfigure}
\hfill
\hspace{0.24\textwidth}
%\\
%\begin{subfigure}{0.24\textwidth}
%\centering
%\includegraphics[width=2cm]{fashionMNIST_3}%
%\caption{$f_2$}
%\label{fig:FashionMNIST:8}
%\end{subfigure}
%\hfil
%\begin{subfigure}{0.24\textwidth}
%\centering
%\includegraphics[width=2cm]{fashionMNIST_dark_3}%
%\caption{$f_2 \LP 100$}
%\label{fig:FashionMNIST:9}
%\end{subfigure}
%\hfil
%\begin{subfigure}{0.24\textwidth}
%\centering
%\includegraphics[width=2cm]{fashionMNIST_light_3}%
%\caption{$f_2 \LM 100$}
%\label{fig:FashionMNIST:10}
%\end{subfigure}
%\hfil
%\begin{subfigure}{0.24\textwidth}
%\centering
%\includegraphics[width=2cm]{aspDist_GT_3}%
%\caption{$g =Asp_{b_r}^{\protect \LP}f_2$}
%\label{fig:FashionMNIST:11}
%\end{subfigure}
%\\
%\begin{subfigure}{0.24\textwidth}
%\centering
%\includegraphics[width=2cm]{aspDist_pred_3}%
%\caption{$Asp_{b}^{\protect \LP}f_2$\\\hfil}
%\label{fig:FashionMNIST:12}
%\end{subfigure}
%\hfil
%\begin{subfigure}{0.24\textwidth}
%\centering
%\includegraphics[width=2cm]{aspDist_dark_pred_3}%
%\caption{$Asp_{b}^{\protect \LP}(f_2 \protect \LP 100)$}
%\label{fig:FashionMNIST:13}
%\end{subfigure}
%\hfil
%\begin{subfigure}{0.24\textwidth}
%\centering
%\includegraphics[width=2cm]{aspDist_light_pred_3}%
%\caption{$Asp_{b}^{\protect \LP}(f_2 \protect \LM 100)$}
%\label{fig:FashionMNIST:14}
%\end{subfigure}
%\hfil
%\hspace{0.24\textwidth}
\caption{(a) Image $f$ coming from the Fashion MNIST test dataset.
(b) $f \protect \LP 100$  darkened image.
(c) $f \protect \LM 100$ brightened image.
(d) $Asp_{b_r}^{\protect \LP}f$: ground-truth $g$, i.e. map of Asplund distances with the reference probe $b_r$.
Predictions $\widehat{g}$, i.e. maps of Asplund distances with the learnt probe $b$ : (e)
in the initial image $Asp_{b}^{\protect \LP}f$ ; (f) 
in the darkened image $Asp_{b}^{\protect \LP}(f \protect \LP 100)$; (g) in the brightened image $Asp_{b}^{\protect \LP}(f \protect \LM 100)$.}
\label{fig:FashionMNIST}
\end{figure}

\begin{figure}[hbt!]
\centering
\begin{subfigure}{0.24\textwidth}
\centering
\includegraphics[width=2.5cm]{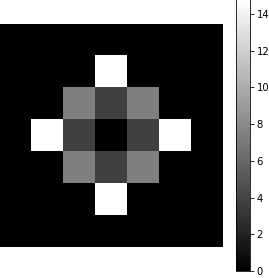}%
\caption{$W_{h,b_r}$}
\label{fig:kernels:1}
\end{subfigure}
\hfill
\begin{subfigure}{0.24\textwidth}
\centering
\includegraphics[width=2.5cm]{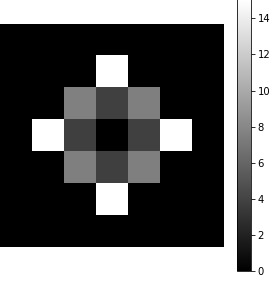}
\caption{$W_{h}$}
\label{fig:kernels:2}
\end{subfigure}
\hfill
\begin{subfigure}{0.24\textwidth}
\centering
\vspace{0.75em}\includegraphics[width=2cm]{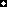}\vspace{0.75em}%
\caption{$W_{m,b_r}$}
\label{fig:kernels:3}
\end{subfigure}
\hfill
\begin{subfigure}{0.24\textwidth}
\centering
\includegraphics[width=2.5cm]{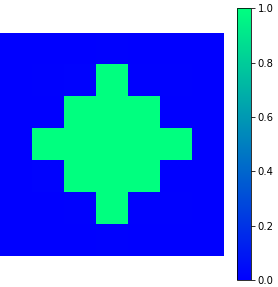}%
\caption{$W_{m}$}
\label{fig:kernels:4}
\end{subfigure}
\caption{In the inverted grey scale, height kernels (a) $W_{h,b_r}$ of the reference probe $b_r$ and (b) $W_h$ of the learnt probe $b$. Mask kernels (c) $W_{m,b_r}$ of the reference probe $b_r$ and (d) $W_{m}$ of the learnt probe $b$. As $W_m$ is a soft mask, a color scale was used.}
\label{fig:kernels}
\end{figure}

Figure~\ref{fig:FashionMNIST} shows an image $f$ of the test set  (\figurename~\ref{fig:FashionMNIST:1}). This image was darkened $f \LP 100$ (\figurename~\ref{fig:FashionMNIST:2}) or brightened $f \LM 100$ (\figurename~\ref{fig:FashionMNIST:3}). The ground-truth $g$ was computed with the map of Asplund distances with the reference probe $b_r$: $g= Asp_{b_r}^{\protect \LP}f$ (\figurename~\ref{fig:FashionMNIST:4}). The predictions $\widehat{g}$ of the maps of Asplund distances were made with the learnt structuring function $b$:
\begin{inparaenum}[(i)] 
\item in the initial image:
$Asp_{b}^{\protect \LP}f_1$ (\figurename~\ref{fig:FashionMNIST:5}), 
\item in the darkened images: $Asp_{b}^{\protect \LP}(f_1 \protect \LP 100)$ (\figurename~\ref{fig:FashionMNIST:6}) 
and 
\item in the brightened image: $Asp_{b}^{\protect \LP}(f_1 \protect \LM 100)$ (\figurename~\ref{fig:FashionMNIST:7}). 
\end{inparaenum}
These images show that there is no noticeable differences between the predictions $Asp_{b}^{\protect \LP}f$, $Asp_{b}^{\protect \LP}(f \protect \LP 100)$, $Asp_{b}^{\protect \LP}(f \protect \LM 100)$ and the ground-truth $g= Asp_{b_r}^{\protect \LP}f$ and that the predictions are robust to lighting variations which are modelled by the LIP-addition or the LIP-subtraction of a constant.
Figure~\ref{fig:kernels} shows that the height kernels $W_{h,b_r}$ of the reference probe $b_r$ and $W_h$ of the learnt probe $b$ are similar. The mask kernels $W_{m,b_r}$ and $W_m$ are also similar. 
%\textcolor{red}{Ajouter des images des fonctions structurantes et de la base de données}.

% ------------------- Figure MNIST dataset

% ------------------- Figure probe kernels

%\FloatBarrier
\vspace{4em}
%%%%%%%%%%%%%%%%%%%%%%%%%%%%%%%%%%%%%%%%%%%%%%%%%%%%%%%%%%%%%%%%%%%%%%%%%%
%
%		Conclusion
%
%%%%%%%%%%%%%%%%%%%%%%%%%%%%%%%%%%%%%%%%%%%%%%%%%%%%%%%%%%%%%%%%%%%%%%%%%%
\section{Conclusion}
\label{sec:concl}

We have introduced a logarithmic morphological neural network which is robust to real lighting variations. Those variations are modelled by the LIP-addition of a constant and they are caused by a change in the light intensity or in the camera exposure-time. Such a neural net is based on the functional Asplund metric defined with the LIP-additive law. In the future, by combining several logarithmic morphological layers, we will define neural nets for numerous practical applications where the light is uncontrolled. We will also study neural nets for the LIP-multiplicative Asplund metric, which is invariant under changes of opacity. Such changes are modelled by the LIP-multiplication by a scalar.

\bibliographystyle{splncs04}
\bibliography{refs}

%Author, F.: Article title. Journal \textbf{2}(5), 99--110 (2016)
%Author, F., Author, S.: Title of a proceedings paper. In: Editor,
%F., Editor, S. (eds.) CONFERENCE 2016, LNCS, vol. 9999, pp. 1--13.
%Springer, Heidelberg (2016). \doi{10.10007/1234567890}
%Author, F., Author, S., Author, T.: Book title. 2nd edn. Publisher,
%Author, A.-B.: Contribution title. In: 9th International Proceedings
%on Proceedings, pp. 1--2. Publisher, Location (2010)
%LNCS Homepage, \url{http://www.springer.com/lncs}. Last accessed 4
%\end{thebibliography}
\end{document}

%% file: header.tex
\usepackage[utf8]{inputenc} % as usual
\usepackage[T1]{fontenc}
\usepackage{graphicx} % required to import graphic files
\graphicspath{{./Fig/}}
\DeclareGraphicsExtensions{.pdf,.jpeg,.png,.jpg}

\usepackage{color} %allows to mark some entries in the tables with color
\usepackage{everyshi} % these two are required to add the little picture on top of every page
\usepackage{placeins} % pour focer l'affichage des figures avant \FloatBarrier

\usepackage{caption}
\usepackage{subcaption}

\usepackage[pdftex,hidelinks,breaklinks]{hyperref}% this enables jumping from a reference and table of content in the pdf file to its target
\hypersetup{     
    urlcolor=cyan,
}
\usepackage{doi}
\usepackage[english]{babel}% to support underscores
\usepackage[strings]{underscore} % for underscores in doi

\usepackage{paralist}

\usepackage{amsmath}
\usepackage{amsfonts}
\usepackage{amssymb}
\usepackage{bm}
\usepackage{algorithmicx}
\usepackage{algpseudocode}
\usepackage{algorithm}
\usepackage{mathrsfs}
\usepackage{scalerel}

\usepackage{etoolbox}
\newcommand{\zerodisplayskips}{%
  \setlength{\abovedisplayskip}{1pt}%
  \setlength{\belowdisplayskip}{1pt}%
  \setlength{\abovedisplayshortskip}{1pt}%
  \setlength{\belowdisplayshortskip}{1pt}}
\appto{\normalsize}{\zerodisplayskips}
\appto{\small}{\zerodisplayskips}
\appto{\footnotesize}{\zerodisplayskips}

\BeforeBeginEnvironment{figure}{\vskip-3ex}
\AfterEndEnvironment{figure}{\vskip-3ex}

\usepackage{array}

\DeclareUnicodeCharacter{0302}{*************************************}

%% file: header_math_commands.tex
\newcommand\scale[2]{\vstretch{#1}{\hstretch{#1}{#2}}}
	
\newcommand{\LP}{
		\mathbin{\mathpalette\LIPcls+}
	}
\newcommand{\LM}{
		\mathbin{\mathpalette\LIPcls-}
		}
\newcommand{\LT}{
		\mathbin{\mathpalette\LIPcls\times}
	}
\newcommand{\LIPcls}[2]{%
		\ooalign{$#1\bigtriangleup$\crcr  \hidewidth\raisefix{#1}\hbox{$#1\scale{0.45}{\bm{#2}}$}\hidewidth}}
\def\raisefix#1{%
  \ifx#1\displaystyle
    \raise.14em
  \else
    \ifx#1\textstyle
      \raise.14em
    \else
      \ifx#1\scriptstyle
        \raise.112em
      \else
        \raise.0933em
      \fi
    \fi
  \fi
}

\makeatletter
\DeclareRobustCommand\bigop[1]{%
  \mathop{\vphantom{\sum}\mathpalette\bigop@{#1}}\slimits@
}
\newcommand{\bigop@}[2]{%
  \vcenter{%
    \sbox\z@{$#1\sum$}%
    \hbox{\resizebox{\ifx#1\displaystyle1.1\fi\dimexpr\ht\z@+\dp\z@}{!}{$\m@th#2$}}%
  }%
}
\makeatother

\newcommand{\Real}{\mathbb R}
\newcommand{\Realb}{\overline{\Real}}

\newcommand{\Zint}{\mathbb Z}
\newcommand{\Ind}{\mathbf{1}}% requires \usepackage{bbm}

\newcommand{\la}{\lambda}

\newcommand{\I}{\mathcal{I}}

\newcommand{\Fcurv}{\mathcal{F}}
\newcommand{\Fcurvb}{\overline{\Fcurv}}

\usepackage{siunitx}
\usepackage{etoolbox}
\newrobustcmd\Bfs{\DeclareFontSeriesDefault[rm]{bf}{b}\bfseries} % pour \'ecrire en gras
\robustify\bfseries
\newcommand\mcc[1]{\multicolumn{1}{c}{#1}} % pour les lignes de textes dans les colonnes de chiffres S